\documentclass[10.7pt,]{article}
\PassOptionsToPackage{prologue,dvipsnames,svgnames}{xcolor}
\usepackage{mdframed} 
\usepackage{soulutf8} 
\usepackage{soul} 

\newcommand{\hlyellow}[1]{{\sethlcolor{yellow}\hl{#1}}}
\newcommand{\hlpink}[1]{{\sethlcolor{pink}\hl{#1}}}
\usepackage[letterpaper, margin=2.54cm, top=2.54cm]{geometry}
\usepackage[super,comma,sort&compress]{natbib}
\usepackage{lmodern}
\usepackage{wrapfig}
\usepackage{authblk} 
\usepackage{amssymb,amsmath}
\usepackage{wrapfig}
\usepackage{graphicx,grffile}
\usepackage[labelfont=bf,labelsep=period]{caption}
\usepackage{ifxetex,ifluatex}
\usepackage{fixltx2e} 
\usepackage{enumitem}
\usepackage[breakable]{tcolorbox}
\usepackage{marvosym}

\ifnum 0\ifxetex 1\fi\ifluatex 1\fi=0 
  \usepackage[T1]{fontenc}
  \usepackage[utf8]{inputenc}
\else 
  \ifxetex
    \usepackage{mathspec}
  \else
    \usepackage{fontspec}
  \fi
  \defaultfontfeatures{Ligatures=TeX,Scale=MatchLowercase}
\fi
\IfFileExists{upquote.sty}{\usepackage{upquote}}{}
\IfFileExists{microtype.sty}{%
	\usepackage{microtype}
	\UseMicrotypeSet[protrusion]{basicmath} 
}{}

\usepackage{listings}
\usepackage{tabularx, booktabs}
\usepackage{multirow}
\usepackage{xcolor}

\usepackage{sectsty}
\allsectionsfont{\fontsize{10}{10}\selectfont}
\subsectionfont{\itshape\bfseries\fontsize{10}{10}\selectfont}
\subsubsectionfont{\normalfont\itshape}

\makeatletter
\renewcommand\subsubsection{\@startsection{subsubsection}{3}{\z@}%
	{-3.25ex\@plus -1ex \@minus -.2ex}%
    {-1.5ex \@plus -.2ex}
    {\normalfont\itshape}}
\makeatother

\makeatletter 
\renewcommand\@biblabel[1]{#1.} 
\makeatother %

\usepackage{etoolbox}
\makeatletter
\patchcmd{\@maketitle}{\LARGE}{\bfseries\fontsize{15}{16}\selectfont}{}{}
\makeatother

\makeatletter
\def\maxwidth{\ifdim\Gin@nat@width>\linewidth\linewidth\else\Gin@nat@width\fi}
\def\maxheight{\ifdim\Gin@nat@height>\textheight\textheight\else\Gin@nat@height\fi}
\makeatother

\setkeys{Gin}{width=\maxwidth,height=\maxheight,keepaspectratio}
\ifx\paragraph\undefined\else
\let\oldparagraph\paragraph
\renewcommand{\paragraph}[1]{\oldparagraph{#1}\mbox{}}
\fi
\ifx\subparagraph\undefined\else
\let\oldsubparagraph\subparagraph
\renewcommand{\subparagraph}[1]{\oldsubparagraph{#1}\mbox{}}
\fi
\usepackage{hyperref}
\hypersetup{
	unicode=true,
	pdftitle={My Cool Title Here},
	pdfauthor={Author One, Author Two, Author Three},
	pdfkeywords={keyword1, keyword2},
	pdfborder={0 0 0},
	breaklinks=true
}
\title{\vspace{-2em} MKRAG: Medical Knowledge Retrieval Augmented Generation for Medical Question Answering}
\author[ ]{\bf\fontsize{13}{14}\selectfont Yucheng Shi\textsuperscript{1,2}\footnote{Three authors contributed equally to this paper. Correspondence: Xiang Li (xli60@mgh.harvard.edu) and Ninghao Liu (ninghao.liu@uga.edu).}, Shaochen Xu\textsuperscript{1*}, Tianze Yang\textsuperscript{1*}, Zhengliang Liu\textsuperscript{1}, Tianming Liu\textsuperscript{1}, Quanzheng Li\textsuperscript{2}, Xiang Li\textsuperscript{2\Letter}, Ninghao Liu\textsuperscript{1\Letter}\vspace{-.7em}}
\affil[1]{\bf\fontsize{13}{14}\selectfont School of Computing, University of Georgia, Athens, GA 30602 USA;}
\affil[2]{\bf\fontsize{13}{14}\selectfont Department of Radiology, Massachusetts General Hospital and Harvard Medical School, Boston, MA 02114, USA}
\date{} 

\begin{document}
\maketitle
\vspace{-4em} 

\section{Abstract}\label{abstract}
\emph{Large Language Models (LLMs), although powerful in general domains, often perform poorly on domain-specific tasks such as medical question answering (QA). In addition, LLMs tend to function as "black-boxes", making it challenging to modify their behavior. 
To address the problem, our work employs a transparent process of retrieval augmented generation (RAG), aiming to improve LLM responses without the need for fine-tuning or retraining. Specifically, we propose a comprehensive retrieval strategy to extract medical facts from an external knowledge base, and then inject them into the LLM's query prompt. Focusing on medical QA, we evaluate the impact of different retrieval models and the number of facts on LLM performance using the MedQA-SMILE dataset. 
Notably, our retrieval-augmented Vicuna-7B model exhibited an accuracy improvement from 44.46\% to 48.54\%. This work underscores the potential of RAG to enhance LLM performance, offering a practical approach to mitigate the challenges posed by black-box LLMs.}

\section{Introduction}
Large Language Models (LLMs) have achieved state-of-the-art performance on a variety of tasks such as commonsense question-answering, translation, and text generation due to their deep architectures and vast number of parameters~\cite{vaswani2023attention}. They excel in comprehending and generating human-like texts due to their training in diverse and extensive text collections~\cite{brown2020language, devlin2018bert}. 

\begin{wrapfigure}{r}{0.35\textwidth}
    \centering
    \includegraphics[width=0.32\textwidth]{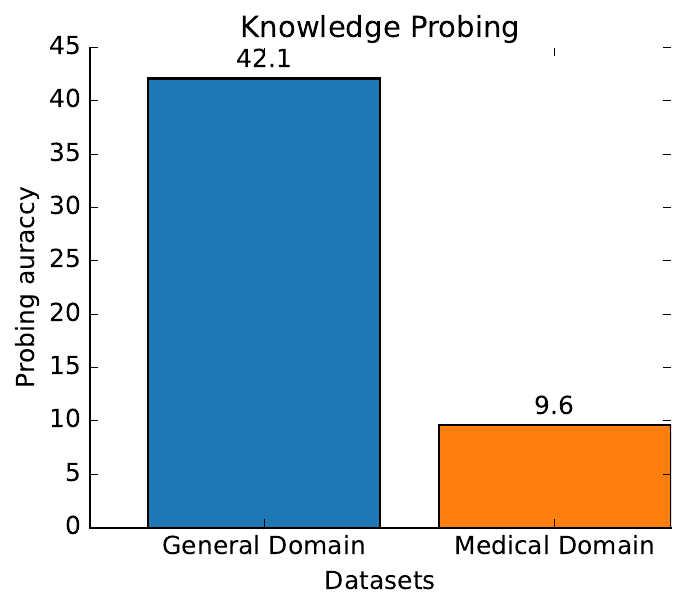}
    \caption{Preliminary Experiment on Knowledge Evaluation: Vicuna-7B demonstrates stronger memorization of general domain knowledge than medical knowledge.}
    \vspace{-5pt}
    \label{pre}
\end{wrapfigure}
However, large language models may lack specific medical knowledge, as this information is usually stored separately and unavailable during model pre-training. To evaluate the quality of medical knowledge encoded in LLMs, we conduct a preliminary experiment as shown in Figure~\ref{pre}. In this experiment, we select Vicuna-7B, a fine-tuned LLaMA-7B~\cite{chiang2023vicuna}, as our candidate large language model, and a reconstructed Disease Database~\cite{yasunaga2021qa} which contains 44,561 triplets as our medical knowledge base. Each triplet is represented in the format $(head, relation, tail)$. We randomly selected 1,000 facts (triplets). For each fact, we prompted Vicuna-7B to infer the tail entity based on the given head entity and relation. 
If Vicuna-7B's answer contained the correct tail entity, we consider that Vicuna-7B has encoded this medical fact correctly. Otherwise, it indicates a failure to encode the relevant medical information. For comparison, we also evaluated LLM performance on the CounterFact dataset~\cite{meng2022locating}, which consists of general domain factual knowledge. The results revealed that Vicuna performed relatively poorly in answering medical knowledge questions but achieved much better performance in the general knowledge domain. This discrepancy highlights the challenges in medical knowledge understanding for LLMs and underscores the need for further knowledge retrieval.

To address the above problem, we propose to conduct medical knowledge retrieval~\cite{shi2024retrieval, wu2024usable}, which refers to the process of retrieving specific knowledge to improve LLM's performance on medical question-answering tasks. Through knowledge retrieval, we attempt to add medical knowledge to LLMs to enhance their ability to understand, answer, or generate content related to medical queries.


Integrating retrieved medical knowledge into LLMs like ChatGPT~\cite{radford2019language} and LLaMA~\cite{touvron2023llama} presents significant challenges. The intrinsic opacity of LLMs such as ChatGPT and the substantial computational costs associated with fine-tuning open-source models like LLaMA limit their capacity for learning the external retrieved knowledge. Given these obstacles, an approach for external knowledge injection becomes imperative. We advocate the use of \textit{in-context learning}, which introduces knowledge into LLMs via prompts~\cite{dong2022survey}. This technique bypasses the complexities of modifying internal model parameters and avoids the need for extensive retraining, facilitating an effective and practical method for infusing external knowledge into LLMs.

However, in-context learning also has limitations, primarily due to the restricted input context length of LLMs, prompting us to carefully select relevant knowledge for optimal prompt design. Na\"ive search strategies fail in the medical domain for two reasons. Firstly, entity matching becomes especially challenging due to the numerous aliases for medical terms. Secondly, a direct embedding search strategy, such as simultaneously embedding both questions and answer candidates and then searching, can lead to misleading retrieval. This is because questions generally contain richer contextual information compared to answers, potentially causing essential answer-related facts to be overlooked in this approach.

In this work, we develop a novel approach to enhance medical knowledge retrieval in language models. (1)~We utilize in-context learning as an innovative mechanism to conduct knowledge injection. By directly combining retrieved knowledge with the model's input context, rather than relying on intricate fine-tuning and resource-intensive retraining processes, we achieve substantial performance improvements in medical QA tasks. (2)~We introduce a tailored fact extraction strategy specifically designed for medical QA. By employing a two-step search process, our approach ensures the retrieval of the most crucial and contextually relevant information from a large external knowledge base.

\section{Related Work}
\subsection{Medical QA \& Large Language Model}
LLMs have been extensively evaluated in specialized medical domains. A Mayo Clinic study found GPT-4 excelled in answering complex radiation oncology physics questions, especially when prompted to explain before answering, showcasing its versatility in complex tasks~\cite{holmes2023evaluating}. Researchers assessed 32 LLMs for interpreting radiology reports, underscoring their varied capabilities in medical NLP~\cite{liu2023evaluating}. A comprehensive review highlighted the diverse evaluation methods for LLMs, emphasizing their significance in medical, ethical, and educational applications~\cite{chang2023survey}. This demonstrates the critical role of LLMs in enhancing medical QA systems and delivering valuable insights.

\subsection{Retrieval Method}
Retrieval methods enhance various applications, including ODQA, where the Retrieval Augmented Generation (RAG) model initially used a Wikipedia-based knowledge base. However, it struggled in domain-specific areas like healthcare. The proposed~\textit{RAG-end2end} addresses this by jointly training the retriever and generator with domain-specific knowledge bases, ensuring all components are updated during training~\cite{siriwardhana2022improving}. Similarly, Retrieval-Augmented Language Modeling (RALM) uses grounding documents during generation, with In-Context RALM showing significant performance gains without modifying the original LM architecture~\cite{ram2023incontext}. In medicine, the Almanac model improved factuality in clinical scenarios by leveraging medical guidelines, supporting clinical decision-making~\cite{hiesinger2023almanac}. ChatGPT’s retrieval feature, still in beta, combines literature search with LLMs for enhanced information retrieval in medical contexts~\cite{jin2023retrieve}. Another concurrent work also shows RAG helps medical question answering~\cite{xiong2024benchmarking}. These retrieval techniques, when integrated with LLMs, offer promising advancements in medical information retrieval and effectiveness across various domains.

\section{Methodology}
\begin{figure*}[t]
    \centering
    \includegraphics[width=1\textwidth]{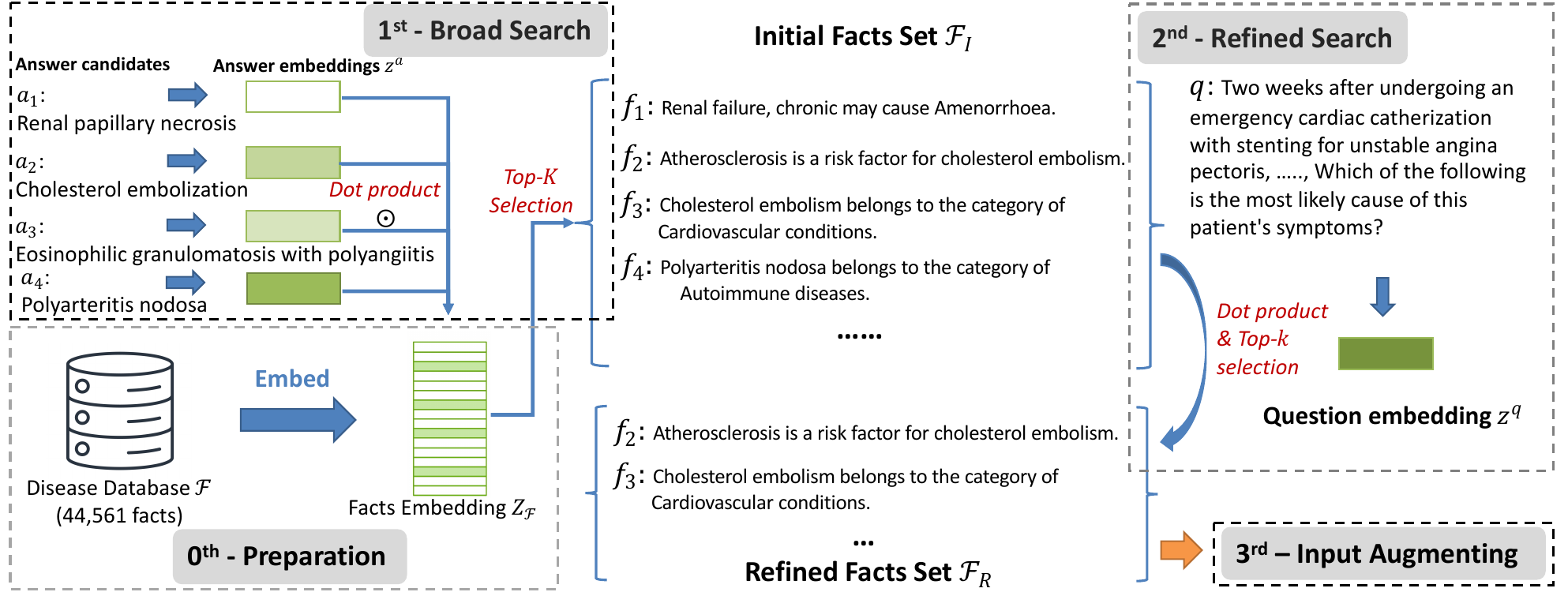}
    \caption{Framework Design for Our Proposed MKRAG: (0) Preparation: use an embedding model to convert fact triplets into embedding. (1) Broad Search: search the most related facts to answer candidates and form the initial facts set. (2) Refined Search: select facts related to the question from the initial facts set and form the refined facts set. (3) Apply retrieved facts to in-context augmenting.}
    \label{fig2}
\end{figure*}

In this study, we propose a method for retrieving and integrating medical knowledge into language models to enhance its ability to answer medical queries. We will incorporate external medical knowledge into the model's prompts to facilitate more accurate reasoning. Initially, we will outline our proposed medical retrieval-augmented generation method. Then, we will detail our strategy for medical knowledge retrieval. Finally, we will discuss our design for knowledge injection. The whole framework is shown in Figure~\ref{fig2}.

\subsection{MKRAG for Medical QA}
We propose to utilize a medical retrieval method to boost the performance of LLMs in answering medical questions (Medical QA tasks). We focus on a typical Medical QA scenario in which each question is accompanied by several answer choices~\cite{jin2021disease}. To correctly answer such a question, one must select the appropriate response from these options. 

Formally, for one medical question $q$, there are four candidate answers $a_1, a_2, a_3,$ and $a_4$. A language model $g_{\theta}(\cdot)$ is expected to select the correct answer $a^*$ from the four candidates given the question $q$ and a question prompt template $t_q$. This process can be formulated as $g_{\theta}(t_q, q) = a^*$. An example question template for a medical QA task is shown below:
\begin{lstlisting}
Given question: [q], which of the following answers is true: [a_1], [a_2], [a_3],
[a_4]. You can only output the predicted label in exact words. No other words
should be included.
\end{lstlisting}
In this paper, MKRAG is introduced as a solution aimed at augmenting language models with medical knowledge, which may either be missing or inaccurately represented in traditional methods. MKRAG is designed with two key phases: (1) the acquisition of medical facts, and (2) the subsequent injection of this knowledge into the model. The first phase, \textit{Medical Facts Retrieval}, is dedicated to identifying and collecting crucial medical information that has the potential to improve the model's ability to respond to the specific medical question. However, merely collecting medical facts is not enough. Thus, the \textit{Knowledge Injection} phase is included to integrate knowledge into LLMs, which will enable language models to understand and utilize the gathered information effectively in their decision-making processes.

Ideally, we would prefer to incorporate all knowledge from the external knowledge base to ensure comprehensiveness. However, we must limit the number of facts considered due to input length constraints. Moreover, redundant information in the model input could also degrade the QA task performance~\cite{cuconasu2024power}. Therefore, the key challenge is to retrieve knowledge closely aligned with the question and useful for finding the correct answer. Our approach focuses on retrieving the most relevant knowledge for question $q$ and answer candidates $a$, ensuring it guides the model toward the correct answer. In the next section, we will discuss how to efficiently retrieve and integrate highly relevant knowledge into LLMs for medical questions.

\subsection{Medical Facts Retrieval}
Formally, we define our retrieval objective as follows: Given a question $q$ and four answer candidates, $a_1, a_2, a_3, a_4$, we aim to identify and extract a set of facts $\{f_1, f_2,...,f_n\}$ that possess the highest relevance to both the question and the answer candidates from an external knowledge base $\mathcal{F}$. Subsequently, we employ these extracted facts as the prompt to conduct in-context model augmenting.

To achieve this goal, we introduce a strategy for comprehensively extracting relevant facts. In the initial preparatory phase, we transform the entire external knowledge base into embeddings for dense retrieval. Specifically, for each fact $f_i$ within the knowledge base $\mathcal{F}$, we employ a pre-trained language model $g_z$ to convert it into an embedding denoted as $z^f_i$. This process results in the creation of an embedded knowledge base, represented as $Z_{\mathcal{F}}$. In our research, we select the Disease Database~\cite{yasunaga2021qa} as our knowledge base $\mathcal{F}$ and employ various models, including SapBert~\cite{liu2020self} and Contriver~\cite{izacard2021unsupervised}, as the embedding model $g_z$. The rationale for using dense retrieval over sparse retrieval methods, like keyword matching, is that medical symptoms and descriptions often vary in their expressions, making them challenging to capture with sparse retrieval techniques.

Following this, in the facts extraction step, we employ the same embedding techniques to convert each answer candidate $a_i$ into an embedding $z^a_i$. For every candidate answer, we then extract the $K$ most closely related facts from the external knowledge base $\mathcal{F}$, establishing an initial set of facts denoted as $\mathcal{F}_I$. In this work, the semantic relevance is measured by the embeddings similarity $s$, which is defined as below:
\begin{equation}
    s(z^a, z^f) = (z^a)^{T} \cdot z^f.
\end{equation}
For  $z^a$, the top-$K$ most related fact set can be selected : 
\begin{equation}
    \mathcal{F}_I = \underset{{f \in \mathcal{F}}}{\text{Top-}K} \,\, s(g_z(f), z^a),
\end{equation}
where the $\text{Top-}K$ function returns $K$ facts with the highest similarity value. The initial set comprises medical information pertaining to all answer candidates, which could be integrated into the augmenting prompt to aid the language model in its reasoning process. However, the retrieved set of facts $\mathcal{F_I}$ may include redundant information that is unrelated to the question description $q$. The inclusion of irrelevant information could potentially confuse the language model, resulting in a decrease in answering performance~\cite{shi2023large}.

Thus, we need to remove these redundant facts, which leads to our fact refinement in the second step of the extraction. Here, we first convert the question $q$ concatenated with four candidates into an embedding $z^q$ using the same model $g_z$. Subsequently, we select the top-$k$ facts from the initial facts set $\mathcal{F}_I$ that exhibit high similarity to $z^q$, forming the refined facts set $\mathcal{F}_R$, which can be defined as below:
\begin{equation}
    \mathcal{F}_R = \underset{{f \in \mathcal{F}_I}}{\text{Top-}k} \,\, s(g_z(f), z^q).
\end{equation}
The refined facts set $\mathcal{F}_R$ contains facts that are related to both question description and answer candidates. 
These contextual facts will act as anchors, providing the model with vital background information intended to improve its decision-making capability. Our hypothesis is that these contextual facts will aid the model in better understanding and aligning its responses with the 
the specific medical question at hand, ultimately enhancing accuracy.

\subsection{Knowledge Injection}
Following the retrieval of medical facts, our methodology progresses to the phase of knowledge injection. This paper employs in-context learning for this purpose, which capitalizes on the inherent abilities of language models to internalize and apply the newly incorporated medical knowledge, thus significantly enhancing their functionality.
Specifically, we directly incorporate the retrieved medical knowledge into the question prompt.  Successful MKRAG can effectively calibrate the output of a pre-trained language model, thereby enhancing its performance on medical question-answering datasets.

In our case, the retrieved medical fact set is $\mathcal{F}_R$, which comprises multiple facts $\mathcal{F}_R = \{f_1, f_2,...,f_n\}$. Each fact can be denoted as a triple $f = (h, r, t)$, where $h$ signifies the head entity, $r$ denotes the relation, and $t$ represents the tail entity. In the medical context, the fact could be a medical statement, such as \textit{(Atherosclerosis, is a risk factor for, cholesterol embolism)}, which is shown in Figure~\ref{fig2}.
We define the injection template as $t_e$, which can be designed like: 
\begin{lstlisting}
Here are some medical facts: f_1, f_2,...f_n. Given question: [q], which of
the following answers is true: [a_1], [a_2], [a_3], [a_4]. You can only output
the predicted label in exact words. No other words should be included.
\end{lstlisting}
The sole distinction between the original template and injection template,  $ t_q$ and $ t_e$, lies in the inclusion of additional medical facts $ f_1, f_2,..., f_n$. These facts are incorporated into the model input to enrich the medical context from a trusted knowledge base. The LLM then reasons over this information along with the question to generate a well-informed response, effectively combining accurate medical knowledge with the necessary complex reasoning for question answering.

\section{Experiments}
To explore the efficacy of retrieval augmented generation in medical question answering, we conduct experiments driven by three key questions: \textbf{RQ1:} Can medical retrieval enhance performance? \textbf{RQ2:}Which knowledge retrieval model is most effective? \textbf{RQ3:} Does the number of retrieval facts impact performance?

\subsection{Experiment Setting}

In this section, we present our experimental settings, including the test dataset, the target language model for medical retrieval, baseline models for comparison, and our evaluation methodology with defined criteria and metrics.

\subsubsection{MedQA-USMILE Dataset}


In our study, we utilized the MedQA-USMLE dataset, a comprehensive resource tailored for evaluating medical question-answering models. This dataset comprises multiple-choice questions, each offering four potential answers, of which only one is correct. The questions are derived from professional medical exams, including the \textit{United States Medical Licensing Examination} (USMLE), \textit{Mainland China Medical Licensing Examination} (MCMLE), and \textit{Taiwan Medical Licensing Examination} (TWMLE), covering a wide range of medical subjects.
The primary aim of this dataset is to test and drive the development of more advanced open-domain question-answering models. Unlike many existing QA datasets, MedQA-USMLE requires models to retrieve relevant information from extensive medical textbooks and perform complex logical reasoning to arrive at the correct answer. This adds a significant challenge to the QA task. 
The dataset spans three languages English, simplified Chinese, and traditional Chinese. 
In this paper, we specifically utilized the English questions portion of the dataset.

\subsubsection{Language Model: Vicuna-7B}

The model leveraged in our study is Vicuna-7B \cite{chiang2023vicuna}, an innovative open-source chatbot developed by fine-tuning the LLaMA model on an expansive dataset derived from user-shared conversations on ShareGPT. The dataset comprised approximately 70K conversations, ensuring a diverse and rich training set. Modifications were made to the training scheme based on the Stanford Alpaca project~\cite{alpaca}. Noteworthy adjustments include accounting for multi-turn conversations in the training loss and significantly extending the maximum context length from the conventional 512, as seen in Alpaca, to 2048 tokens in Vicuna. Though promising, it is essential to underscore that Vicuna-7B, like other LLMs, has certain limitations, which are considered in our experimental design.

\subsubsection{Baseline Models for Medical QA}

In the landscape of language models, our experiments positioned our approach amidst a collection of innovative models. BERT \cite{devlin2018bert} can be used to capture rich context by examining both preceding and following text. One of its derivatives, BioBERT \cite{lee2020biobert}, was specifically designed for the biomedical domain. By pre-training on biomedical corpora, it adeptly navigated the unique terminologies and structures characteristic of biomedical literature., often surpassing BERT and other models in biomedical text mining tasks. Another noteworthy advancement came from RoBERTa \cite{liu2019roberta}. This model revisited the training scheme of BERT, making optimizations in hyperparameters and underscoring the substantial benefits of parameter tuning. SapBERT \cite{liu2020self} was exceptional with its unique self-alignment mechanism, combined with its capacity to exploit vast biomedical ontologies like UMLS, which made it a robust solution for tasks like medical entity linking. Lastly, QA-GNN \cite{yasunaga2021qa} offered a novel answering method. Integrating insights from pre-trained language models with knowledge graphs demonstrated improved reasoning across various data sources, notably on benchmarks such as MedQA-USMLE. In summary, these models, each with their distinct strengths, provided a comprehensive benchmark for evaluating the performance of our experimental approach.

\subsubsection{Answer Evaluation}

For the assessment of our model's performance, we employed a string-matching approach to quantify the alignment between the model-generated answers and the ground truth. In this context, an answer was deemed correct if the entirety of the ground truth was identifiable within the model's output. For baselines, BERT-based models encode the question and answer choices into a sequence, using the \texttt{[CLS]} token’s embedding to represent it, with a fully connected layer and softmax selecting the highest probability answer. QA-GNN integrates knowledge graph information and a language model, scoring KG node relevance to the question to assist in selecting the correct answer.

\subsection{Main Experiment Results}

\begin{table}[ht]
\centering
\caption{Comparison of MedQA-USMLE (Test) Answering Accuracy.}
\label{medqa}
\begin{tabular}{lcc}
\toprule
\textbf{Method} & \textbf{Accuracy (\%)} \\
\midrule
BERT-base~\cite{devlin2018bert}  & 34.3 \\
BioBERT-base~\cite{lee2020biobert} & 34.1 \\
RoBERTa-large~\cite{liu2019roberta} & 35.0\\
BioBERT-large~\cite{lee2020biobert}  & 36.7 \\ 
SapBERT~\cite{liu2020self}         & 37.2 \\
QA-GNN~\cite{yasunaga2021qa}          & 38.0 \\
\midrule
Vanilla Vicuna~\cite{chiang2023vicuna}         & 44.46 \\
\midrule
MKRAG Vicuna (\textbf{Ours})   & \textbf{48.54} \\
\bottomrule
\end{tabular}
\end{table}

As shown in Table~\ref{medqa}, the Vicuna-7B model achieves the best performance in accuracy on the test split of MedQA-USMLE dataset. We also have the following observation: (1) our medical retrieval method for the Vicuna model yielded a significant improvement in accuracy, achieving 48.54\%. This improvement not only outperformed the baseline, the vanilla Vicuna, by over 4\% points but also surpassed the performances of models like Bio-BERT-large, SapBERT, and QA-GNN.

(2) Our approach is more efficient than baselines. The enhancements in the post-retrieval Vicuna model are realized without resorting to resource-intensive methods like fine-tuning (BioBERT) or the overhead of training an entirely new model from scratch (SapBERT and QA-GNN). This underscores the effectiveness of medical retrieval as a strategy, showcasing that with strategic knowledge retrieval and injection, we can achieve competitive performance improvement without the typically associated computational costs.

\subsection{Ablation Study on Retrieval Model}

\begin{wrapfigure}{r}{0.4\textwidth}
    \centering
    \includegraphics[width=0.35\textwidth]{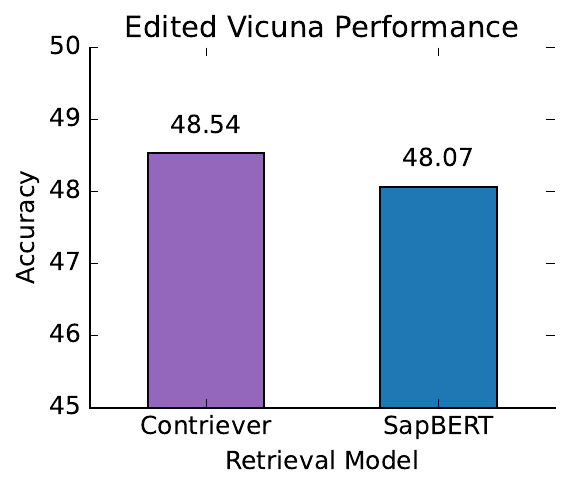}
    \caption{Retrieval Model Comparison: Language model pre-trained on general corpus shows better performance than the model pre-trained on medical corpus.}
    \label{retrievalmodel}
    \vspace{-10pt}
\end{wrapfigure}

In this subsection, we compare the effectiveness of different embedding models in our fact retrieval task. We show the QA accuracy using the Contriever~\cite{izacard2021unsupervised} and SapBert~\cite{liu2020self} as the embedding model $g_z$ in Figure~\ref{retrievalmodel}. The obtained embeddings are used to retrieve related facts, as we discuss in the Medical Facts Retrieval section. We can observe that Contriever slightly outperformed SapBert, securing an accuracy of 48.54\% compared to SapBert's 48.07\%.

The observation that a retriever trained on a general domain outperforms SapBert, a model specialized in domain-specific (medical) datasets, presents an intriguing point for analysis. There are two potential rationales for this phenomenon. First, Contriever utilizes contrastive learning to pre-train the model, which is more effective than self-alignment pretraining~\cite{liu2020self}. Contriever's diverse pre-training corpus, which spans multiple text domains beyond medical texts, endows it with a sophisticated understanding of language semantics and structures. This broad training foundation enables Contriever to more accurately identify relevant facts within a query's context. In contrast, SapBert is primarily designed for specialized medical datasets and is inherently more narrow in its focus. Although both models are used on the same dataset for this experiment, their foundational design principles could result in different retrieval competencies.  

In summary, the results underscore the significance of a retrieval model's architecture and foundational training, even when working within a specialized domain. Contriever's superior accuracy suggests that a broader pre-training approach can offer advantages in specific retrieval tasks, even within a constrained dataset like our disease database.

\subsection{Ablation Study on Retrieval Fact Number}

In the following experiment, we analyze the retrieval performance with different numbers of medical facts. Specifically, we first select $k$ values from $4$, $8$, and $16$, then incorporate the corresponding top-$k$ facts into the prompt. We compare the QA accuracy of the language model in Table~\ref{number}. The results indicate a positive correlation between the number of inserted facts and the model's performance, with a notable improvement as more facts are incorporated.

\begin{wraptable}{r}{0.5\textwidth}
\centering
\caption{Comparison of Retrieval Fact Number.}
\label{number}
\begin{tabular}{lllll}
\toprule
\multirow{2}{*}{Fact Number} & & 4 & 8 & 16 \\ 
 & & \multicolumn{3}{c}{Accuracy (\%)} \\ \hline
\multirow{2}{*}{Retrieval Model} & Contriever &41.86 & 45.48 & 48.54 \\
& SapBert & 41.63 & 45.01 & 48.07 \\
\bottomrule
\end{tabular}
\end{wraptable}

This correlation can be attributed to the nature of medical data. Medicine is a discipline characterized by its intricate web of interrelated facts, pathologies, and treatments. By providing the model with a richer set of facts, we essentially arm it with a more comprehensive context, thereby enabling it to discern finer nuances and interrelations when faced with medical questions. In such a complex domain, every additional piece of relevant information becomes a crucial anchor, aiding in more informed decision-making. However, there's a practical ceiling to this approach, determined by the limitations of the prompt size that can be presented to the model.  Therefore, while the trend suggests that more facts inherently lead to better performance, the real-world application of this insight is bounded by the technical constraints of the language model.

\subsection{Case Study}
\begin{tcolorbox}[breakable, title={Case Study 1: \{Question\_id: 1253\}}, colback=white, colframe=black, colbacktitle=white, coltitle=black, fonttitle=\bfseries, rounded corners]
\textbf{Question:}
A 7-day-old newborn boy presents to the emergency department with a history of fever, irritability, and generalized erythema. During the first 2 days of clinical manifestations, the parents of the child tried to control the symptoms using acetaminophen; however, the newborn continued to be ill, and blisters were noticeable around the buttocks, hands, and feet. During the physical examination, the vital signs include body temperature 39.0°C (102.3°F), heart rate 130/min, and respiratory rate 45/min. Ears, nose, and oral mucosa preserved their integrity, \hlyellow{while the skin presents with diffuse blanching erythema and flaccid blisters with a positive Nikolsky’s sign}. What is the most likely diagnosis for this patient?
\vspace{5pt} %

\textbf{Choices:} 
[A.] Scarlet fever, [B.] Scalded skin syndrome, [C.] Impetigo, [D.] Pyoderma

\vspace{2pt} 

\textbf{Ground Truth: B (Scalded skin syndrome)}
\vspace{2pt} %

\textbf{Retrieved fact:} 

\hlyellow{Staphylococcal scalded skin syndrome may cause Erythema multiforme}, \hlyellow{Staphylococcal scalded skin syndrome belongs to the category of Bacteria and bacterial diseases}, \hlyellow{Staphylococcus aureus may cause Staphylococcal scalded skin syndrome}, Impetigo may cause...

\vspace{2pt} 

\textbf{Answer Before Retrieval:} \\
The most likely diagnosis in this patient is: \hlpink{Scarlet fever}.

\vspace{2pt} 

\textbf{Answer after Retrieval:} \\
The most likely diagnosis in this patient is \hlyellow{Staphylococcal scalded skin syndrome}. The patient's symptoms of fever, irritability, and generalized erythema, along with the presence of blisters on the buttocks, hands, and feet, are consistent with this diagnosis. Additionally, \hlyellow{the patient's vital signs and the physical examination findings of diffuse blanching erythema and flaccid blisters with a positive Nikolsky's sign are also consistent with Staphylococcal scalded skin syndrome}.

\end{tcolorbox}

\vspace{2pt}

\begin{tcolorbox}[breakable, title={Case Study 2: \{Question\_id: 1260\}}, colback=white, colframe=black, colbacktitle=white, coltitle=black, fonttitle=\bfseries, rounded corners]
\textbf{Question:}
A 51-year-old man is brought to the physician by his wife because of a 3-week history of abnormal behavior. His wife reports he has been preoccupied with gambling at the local casino. He becomes restless and irritable when his wife attempts to restrict his gambling. Four months ago, \hlyellow{he was diagnosed with Parkinson disease and treatment was initiated}. Which of the following best describes the mechanism of action of the drug that was most likely prescribed for this patient?

\vspace{2pt} %
\textbf{Choices:} [A.] Peripheral inhibition of DOPA decarboxylase, [B.] Direct activation of dopamine receptors, [C.] Selective inhibition of monoamine oxidase B, [D.] Inhibition of catechol-O-methyl transferase

\vspace{2pt} %
\textbf{Ground Truth: B (Direct activation of dopamine receptors)}
\vspace{2pt} %

\textbf{Retrieved fact:}

Dopamine transporter deficiency syndrome may cause Parkinsonism, \hl{Dopamine may treat Parkinson's disease}, \hl{Dopamine may treat Parkinsonian disorders}, \hl{Dopamine may treat Tremors}, L-DOPA may treat...
\vspace{2pt} 

\textbf{Answer Before Retrieval:} \\
The mechanism of action of the drug most likely prescribed for this patient is: \hlpink{Inhibition of catechol-O-methyl transferase}.

\vspace{2pt} 

\textbf{Answer after Retrieval:} \\
The mechanism of action of the drug most likely prescribed for this patient is: \hlyellow{Direct activation of dopamine receptors}.
\vspace{2pt} 
\end{tcolorbox}

\textbf{Case Study Analysis:} \\
In case study 1, the decision to change the diagnosis from \hlpink{Scarlet fever} to \hlyellow{Staphylococcal scalded skin syndrome} in the \textit{Answer after Retrieval} section was influenced by the retrieved facts highlighting the association of Staphylococcal scalded skin syndrome with erythema multiforme and its causation by Staphylococcus aureus. These highlighted facts directly support the symptoms and physical examination findings described in the question, particularly the presence of blisters and a positive Nikolsky's sign, which are characteristic of Staphylococcal scalded skin syndrome. The inclusion of these specifics from the retrieved information provided a stronger correlation with the patient's condition, leading to the revised diagnosis.

In case study 2, although the retrieved facts do not directly mention the specific mechanism of action of the medication, emphasizing the role of dopamine in treating Parkinson's disease, Parkinsonian disorders, and tremors suggests that direct activation of dopamine receptors might be an effective treatment strategy. This indirect inference from the yellow highlighted information about dopamine's therapeutic role could have prompted the model to shift its decision from \hlpink{Inhibition of catechol-O-methyl transferase} to \hlyellow{Direct activation of dopamine receptors}. This potential rationale for changing the decision based on the retrieved information illustrates how the model might use indirect evidence to optimize its answer strategy.

\section{Conclusion}
In our study, we introduced MKRAG, a method designed to enhance the performance of large language models (LLMs) in answering medical questions. This approach involves medical facts retrieval and knowledge injection. We explored various fact retrieval mechanisms and found that Contriever slightly outperformed SapBert, highlighting the importance of choosing the right retrieval technique. Additionally, we discovered that presenting more facts, within the limits of the prompt, tends to improve performance. 

Our work with MKRAG underscores the potential to enhance language models for specific tasks, such as medical question-answering. These findings not only contribute to our understanding of language model optimization but also suggest pathways for improving LLMs' performance in other domain-specific tasks. 

\section{Acknowledgments}
This work is, in part, supported by NSF (\#IIS-2223768) and Google Research Scholar Program. The views and conclusions in this paper
are those of the authors and should not be interpreted as representing any funding agencies.

\bibliographystyle{vancouver}
\bibliography{amia}

\end{document}